\documentclass[10pt,journal,compsoc]{IEEEtran}
\renewcommand{\baselinestretch}{0.98}
\ifCLASSOPTIONcompsoc
  \usepackage[nocompress]{cite}
\else
  \usepackage{cite}
\fi

\usepackage{graphicx}
\usepackage{comment}
\usepackage{amsmath,amssymb} 
\usepackage[switch]{lineno}
\usepackage{color}
\usepackage{soul}

\usepackage{multirow}
\usepackage{multicol}
\usepackage{floatrow}
\usepackage[noend]{algpseudocode}
\usepackage{algorithmicx,algorithm}
\usepackage{bbding}

\usepackage{tikz,xcolor,hyperref}
\usepackage{subfigure}
\usepackage{booktabs}
\usepackage{makecell}
\usepackage{xspace}

\definecolor{lime}{HTML}{A6CE39}
\DeclareRobustCommand{\orcidicon}{%
    \begin{tikzpicture}
    \draw[lime, fill=lime] (0,0) 
    circle [radius=0.16] 
    node[white] {{\fontfamily{qag}\selectfont \tiny ID}};    \draw[white, fill=white] (-0.0625,0.095) 
    circle [radius=0.007];    \end{tikzpicture}
    \hspace{-2mm}}
\foreach \x in {A, ..., Z}{%
    \expandafter\xdef\csname orcid\x\endcsname{\noexpand\href{https://orcid.org/\csname orcidauthor\x\endcsname}{\noexpand\orcidicon}}
    }

\begin{document}
%
\title{Facial Foundational Model Advances Early Warning of Coronary Artery Disease from Live Videos with DigitalShadow}
%
%
%
%

\author{Juexiao~Zhou$^{1,2,3,\#}$, Zhongyi Han$^{1,2,3,\#}$, Mankun Xin$^{4, \#}$, Xingwei He$^{5, \#}$, Guotao Wang$^{8, \#}$, Jiaoyan Song$^{4}$, Gongning Luo$^{1,2,3}$,  Wenjia He$^{1,2,3}$, Xintong Li$^{4}$, Yuetan Chu$^{1,2,3}$, Juanwen Chen$^{4}$, Bo Wang$^{7}$, Xia Wu$^{4}$, Wenwen Duan$^{4}$, Zhixia Guo$^{4}$, Liyan Bai$^{4}$, Yilin Pan$^{4}$, Xuefei Bi$^{6}$, Lu Liu$^{9}$, Long Feng$^{4}$, Xiaonan He$^{4,*}$, Xin~Gao$^{1,2,3,*}$
\thanks{
$^1$Computer Science Program, Computer, Electrical and Mathematical Sciences and Engineering Division, King Abdullah University of Science and Technology (KAUST), Thuwal 23955-6900, Kingdom of Saudi Arabia.\\
$^2$Center of Excellence for Smart Health, King Abdullah University of Science and Technology (KAUST), Thuwal 23955-6900, Kingdom of Saudi Arabia.\\
$^3$Center of Excellence on Generative AI, King Abdullah University of Science and Technology (KAUST), Thuwal 23955-6900, Kingdom of Saudi Arabia.\\
$^4$Emergency Critical Care Center, Beijing AnZhen Hospital, Affiliated to Capital Medical University, Beijing 100029, China\\
$^5$Tongji Hospital, Wuhan, China\\
$^6$Shanxi Cardiovascular Hospital, China\\
$^7$Beijing Changping Hospital, China\\
$^8$Daqing Longnan Hospital, Daqing, China\\
$^9$Tianjin Academy of Traditional Chinese Medicine Affiliated Hospital\\
$^*$Corresponding author. e-mail: xin.gao@kaust.edu.sa\\
$^\#$These authors contributed equally.
}}

\IEEEtitleabstractindextext{%
\begin{abstract}
Global population aging presents increasing challenges to healthcare systems, with coronary artery disease (CAD) responsible for approximately 17.8 million deaths annually, making it a leading cause of global mortality. As CAD is largely preventable, early detection and proactive management are essential. In this work, we introduce DigitalShadow, an advanced early warning system for CAD, powered by a fine-tuned facial foundation model. The system is pre-trained on 21 million facial images and subsequently fine-tuned into LiveCAD, a specialized CAD risk assessment model trained on 7,004 facial images from 1,751 subjects across four hospitals in China. DigitalShadow functions passively and contactlessly, extracting facial features from live video streams without requiring active user engagement. Integrated with a personalized database, it generates natural language risk reports and individualized health recommendations. With privacy as a core design principle, DigitalShadow supports local deployment to ensure secure handling of user data.
\end{abstract}

}

\maketitle

\IEEEdisplaynontitleabstractindextext

%
\IEEEpeerreviewmaketitle

\section{Introduction}
The world's population is rapidly ageing\cite{chesnaye2024impact}, with significant implications for the prevalence of chronic diseases such as Coronary Artery Disease (CAD)\cite{weintraub2024achieving}, affecting not only individuals but also families and societies at large\cite{Ageing_and_health_2022}. The number of older people is increasing at an unprecedented rate, projected to grow from approximately 761 million in 2021 to 1.6 billion by 2050, which would represent nearly 16\% of the global population, according to the UN's World Social Report 2023\cite{UNDESA_World_Social_Report_2023}. This ageing trend is not limited to developed nations, many developing countries are also experiencing a significant increase in the proportion of older adults\cite{UNDESA_World_Social_Report_2023}. The aging population presents numerous challenges, including increased pressure on healthcare systems, pension schemes, and long-term care facilities, alongside potential economic consequences, which together fuel growing demand for healthcare services\cite{gbd2022global, stennett2022impact}. With advancing age, people become more vulnerable to various critical diseases\cite{de2022late}, such as CAD\cite{savarese2022global}, stroke\cite{diener2022review}, cancer\cite{seale2022making}, and Parkinson's disease (PD)\cite{poewe2017parkinson, tansey2022inflammation, krokidis2022sensor}, leading to considerable morbidity and mortality\cite{shayegan2022real}. Among these, CAD remains the leading cause of death and chronic disability from cardiovascular diseases worldwide. The World Health Organization reports that cardiovascular diseases (CVDs), which include CAD, cause an estimated 32\% of all global deaths\cite{World_Health_Organization}. Luckily, CAD caused by atherosclerosis is largely preventable\cite{kaminsky2022importance}. Therefore, early detection and management of CAD are essential for improving health outcomes and preventing complications, particularly as early identification of at-risk individuals, can greatly reduce disease progression and adverse events\cite{ismail2022psychosis}.

In general, an early warning system in healthcare aims to detect changes in an individual's health status, alerting caregivers and healthcare professionals to take appropriate actions before critical events occur\cite{muralitharan2021machine, meckawy2022effectiveness}. Early warning system for diseases in older adults is especially important for several reasons\cite{oberai2025digital}. First, aging is associated with the decline of the immune system and the increased susceptibility to infections and chronic diseases, increasing health risks for the elderly\cite{mogilenko2022immune}. Second, elderly people often experience multiple comorbidities, making it difficult to detect new disease onsets or complications\cite{van2022challenges}. Third, many elderly people live alone and may not have regular access to healthcare, further complicating early detection of critical conditions\cite{zeng2022analysis}. Given these factors, there is a growing need for an early warning system that can be deployed at home to identify early signs of critical diseases in the elderly, enabling timely interventions and treatments\cite{mcgaughey2007outreach, gao2020machine}.

However, early warning systems for CAD are still in their infancy\cite{chu2023advances}. Recent advances in digital healthcare have made it possible to monitor CAD through ubiquitous intelligent devices\cite{follmer2024roadmap}. Research related to early detection of CAD in home settings is still relatively scarce. Mazumder et al. \cite{mazumder2022synthetic} and Banerjee et al. \cite{banerjee2018time} have demonstrated that irregular or weak pulses can be detected by artificial intelligence (AI) for the assessment of the risk of CAD. Dai et al. used raw heart sound signals for CAD detection as well\cite{dai2024deep}. However, these methods are based on wearable devices and cannot function in a completely contactless manner. Lin et al. \cite{lin2020feasibility} identified significant correlations between CAD risk and facial features, such as earlobe creases, heavy pouches, and nasal folds, and used deep learning (DL) to detect these features automatically. This method can be applied using mobile devices equipped with cameras and computing capabilities, such as smartphones and laptops, to capture facial images and screen for CAD in a contactless and non-intrusive manner. However, its performance is limited by the use of a single image, and it requires active user participation to stay still and collect facial data. Kung et al. \cite{kung2024prediction} predicts CAD based on facial temperature information captured by non-contact infrared thermography. Similarly to the work of Lin et al., this study not only faces performance limitations but also presents challenges for users in obtaining infrared thermography devices. Liu et al. \cite{liu2022videocad} proposed ImageCAD and VideoCAD, which integrate pulse and facial image analysis via video capture to enhance CAD detection. However, these algorithms require users to remain stationary during video recording, limiting their application in dynamic real-world environments. In conclusion, no existing method can continuously predict, track, and manage CAD risk in a completely contactless manner using standard surveillance cameras in real time, 24/7.

Here, we propose DigitalShadow, an early warning system designed for the detection, tracking, and management of CAD through live video streams, enabled by a fine-tuned facial foundation model. DigitalShadow is a fully contactless solution that operates continuously using live video feeds from 24/7 health surveillance cameras. It is well-suited for diverse deployment scenarios, including in-home monitoring for elderly individuals living alone and single-room elderly care settings in healthcare facilities. To develop DigitalShadow, we first pre-trained a facial foundation model on a large-scale dataset of 21 million facial images, capturing a wide range of facial representations and variations. We then fine-tuned this model into a specialized CAD risk assessment model, called LiveCAD, using 7,004 facial images from 1,751 subjects collected across four hospitals. This two-stage training pipeline enables DigitalShadow to robustly extract subtle CAD-related facial features from live video streams. DigitalShadow operates passively, requiring no active engagement from users. It continuously collects facial data from video input, analyzes it in real time, and updates a personalized database containing each individual's CAD risk profile. The system then generates interpretable natural language reports, offering both risk assessments and actionable health recommendations. To ensure privacy and data security, DigitalShadow is designed to support local deployment and edge computing, allowing all data processing to occur on-device or within trusted environments without transmitting sensitive information to the cloud.

\begin{figure*}[htb]
\centering
\includegraphics[width=1\textwidth]{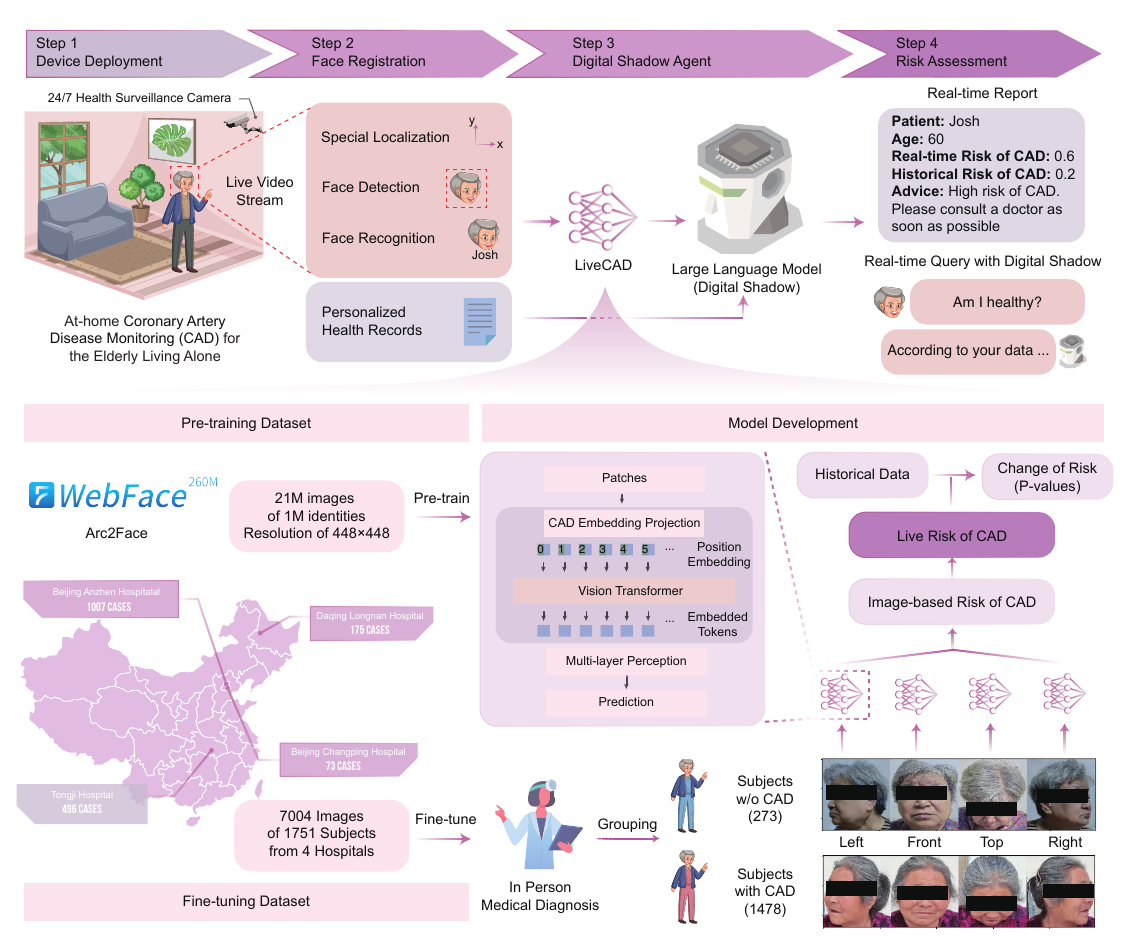}
\caption{\textbf{Overview of the DigitalShadow Solution.} DigitalShadow is a privacy-preserving early warning solution designed for the detection, tracking, and management of CAD using facial imagery, powered by a vertically fine-tuned facial foundation model. The system operates fully contactlessly through continuous video input from 24/7 health surveillance cameras. It performs accurate human localization and face recognition, focusing exclusively on the designated user's face to ensure both analytical precision and data privacy. Each detected face is analyzed by LiveCAD, a specialized model fine-tuned for CAD risk assessment. LiveCAD is built upon a facial foundation model pre-trained on 21 million images from 1 million identities at 448×448 resolution, providing robust general facial representation. This base model is further fine-tuned on an in-house dataset of 7,004 facial images from 1,751 subjects across four hospitals, optimizing it for CAD-related feature extraction and prediction. CAD risk predictions are securely stored in a personalized health record database. Periodic assessments are then interpreted by a LLM, which generates comprehensive, human-readable risk reports. DigitalShadow thus enables real-time, continuous CAD monitoring with high accuracy, user-centric design, and strong privacy guarantees.}
\label{fig1}
\end{figure*}

\section{Results}
\subsection{The overall system design of DigitalShadow}
DigitalShadow is a privacy-preserving early warning system designed for the real-time detection, longitudinal tracking, and proactive management of CAD using live video streams, powered by a fine-tuned facial foundation model (\textbf{Figure} \ref{fig1}). As a fully contactless solution, DigitalShadow operates through continuous video feeds captured by 24/7 health surveillance cameras, making it ideal for passive health monitoring in homes, clinics, and long-term care facilities. The system is capable of accurate human spatial localization, face detection, and facial recognition from live streams. Among all detected faces, only the designated user’s face is processed further, ensuring that bystander or irrelevant facial data is automatically filtered out to protect privacy. Each detected face belonging to the user is then analyzed by LiveCAD, a specialized CAD assessment model fine-tuned from a large-scale, pre-trained facial foundation model (\textbf{Methods}). LiveCAD infers the user's CAD risk level based on facial features extracted from the video. These risk predictions are securely stored and continually integrated into a personalized health record database. Over time, the accumulated CAD risk data is synthesized and interpreted by a large language model (LLM), which generates comprehensive, natural language health reports. These reports include risk summaries and tailored health recommendations, providing both users and clinicians with actionable insights for early intervention. Importantly, DigitalShadow is designed with privacy and data security at its core. All data processing, including inference and report generation, can be performed locally using edge computing, eliminating the need to transmit sensitive data to the cloud. This architecture enables secure, real-time CAD monitoring while maintaining full control over personal health information.

\begin{figure*}[htb]
\centering
\includegraphics[width=0.85\textwidth]{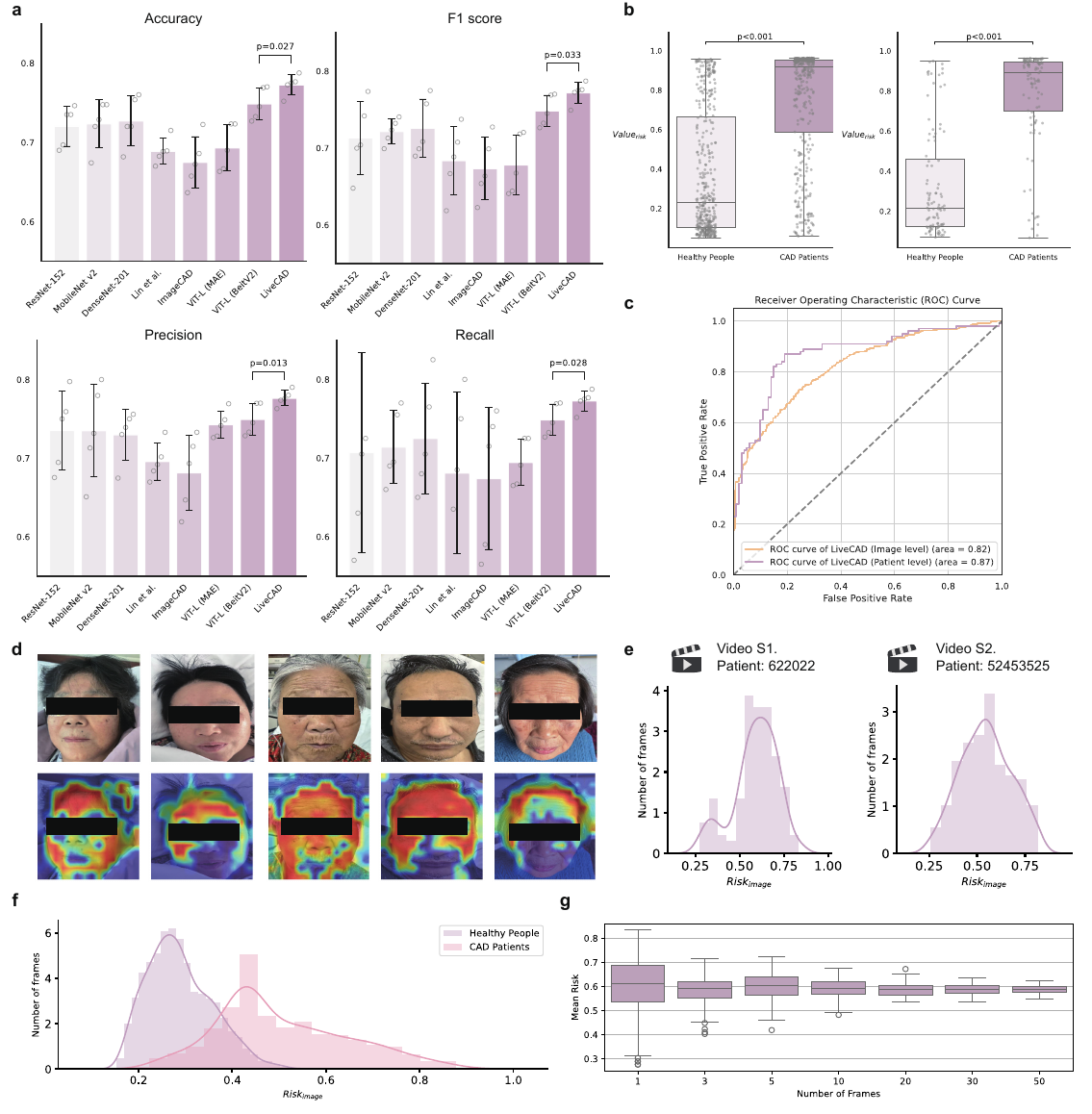}
\caption{\textbf{Evaluation of LiveCAD.} a) Comparison of LiveCAD with baseline methods across key evaluation metrics, including accuracy, F1 score, precision, and recall. b) Boxplot of predicted CAD risk values based on individual facial images (left) and aggregated at the patient level (right). c) Receiver Operating Characteristic (ROC) curves for LiveCAD at both the image (orange color) and the patient (purple color) levels, demonstrating classification performance. d) Visualization of feature representations learned by the trained LiveCAD model, illustrating its ability to distinguish CAD-relevant patterns. e) Temporal distribution of CAD risk values extracted from video streams of two representative patients (IDs: 622022 and 52453525). f) Distribution of CAD risk scores across long-term video data for all CAD patients and healthy subjects, highlighting group-level differences. g) Relationship between observation duration and the robustness of CAD risk predictions. Longer observation times yield more reliable and consistent risk estimates.}
\label{fig2}
\end{figure*}

\begin{figure*}[htb]
\centering
\includegraphics[width=0.8\textwidth]{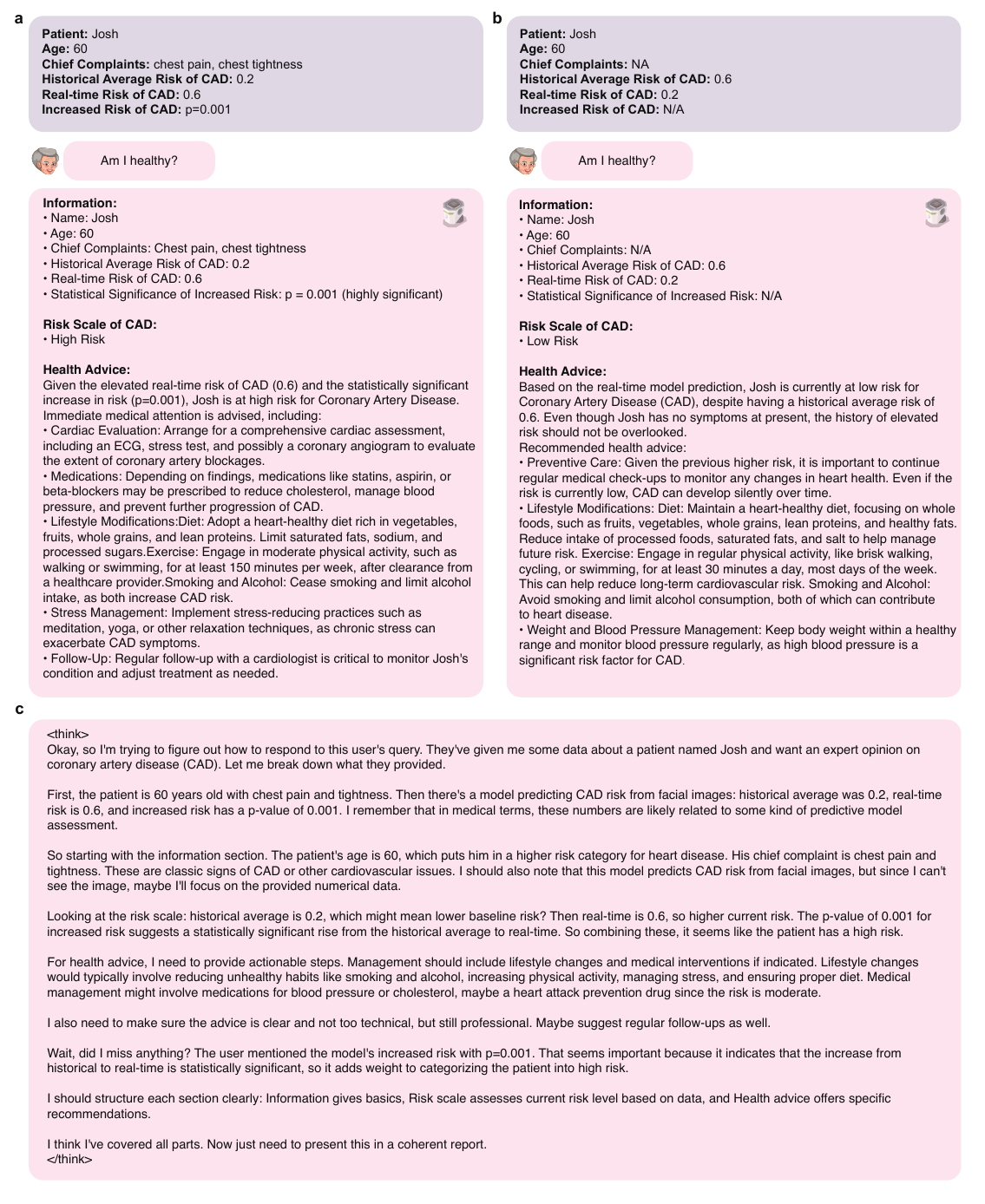}
\caption{\textbf{Health reports generated by DigitalShadow, based on LiveCAD predictions and personalized health records, interpreted through a reasoning LLM.} a) Example of a generated report indicating high CAD risk. b) Example of a generated report indicating low CAD risk. c) Visualization of the reasoning process performed by the LLM for the high-risk case shown in (a), illustrating how risk factors and temporal trends are synthesized into actionable insights. } 
\label{fig3}
\end{figure*}

\subsection{Risk assessment and early warning of CAD from facial images using LiveCAD}
For each facial image extracted from a live video stream, LiveCAD performs an automated assessment of CAD risk. To evaluate its performance, LiveCAD was benchmarked against seven representative methods, encompassing a wide spectrum of model architectures and approaches. These baselines include three classical deep learning models widely used in image analysis tasks, including ResNet-152, MobileNet v2, and DenseNet-201, as well as two state-of-the-art CAD prediction approaches, including the method proposed by Lin et al.\cite{lin2020feasibility}, and ImageCAD\cite{liu2022videocad}. Additionally, to investigate the contribution of LiveCAD’s vertical fine-tuning strategy, we conducted an ablation study using two vision transformer (ViT-L) models pre-trained on ImageNet, including one with Masked Autoencoding (MAE) and the other with BEiT v2 (Methods). 

As illustrated in \textbf{Figure} \ref{fig2}, LiveCAD consistently outperforms all baseline models across four key evaluation metrics: accuracy, F1-score, precision, and recall. Notably, in the ablation study, LiveCAD exhibited statistically significant improvements over the ViT-L model pre-trained with BEiT v2 on ImageNet dataset, with respective P-values of 0.027 (accuracy), 0.023 (F1-score), 0.013 (precision), and 0.028 (recall), demonstrating the efficacy of domain-specific pre-training and fine-tuning for CAD risk prediction from facial images. These results underscore the robustness and clinical relevance of LiveCAD in providing accurate, image-level CAD risk assessments, forming a critical component of the overall DigitalShadow system.

\subsection{Video-Based CAD risk assessment with LiveCAD enhances performance}
While LiveCAD demonstrates strong performance on individual facial images, similar to previous methods such as those proposed by Lin et al. and ImageCAD, relying solely on single-frame analysis can introduce variability and noise, potentially affecting prediction accuracy. As illustrated in Figure \ref{fig2}b, CAD risk values derived from single images exhibit a long-tailed distribution, especially when comparing healthy subjects with CAD patients. This suggests that transient variations in lighting, pose, or expression may compromise the reliability of single-image predictions. To address this limitation, we extended LiveCAD’s inference capabilities to video-based assessments by aggregating predictions from multiple facial images captured across different time points and angles within a video stream. This temporal and multi-view aggregation reduces the influence of outliers and transient noise, resulting in more stable and representative CAD risk profiles. As shown in Figure \ref{fig2}c, this multi-image strategy leads to a 5\% increase in the Area Under the Receiver Operating Characteristic Curve (ROC AUC), demonstrating a notable improvement in both prediction accuracy and robustness. This finding underscores the value of continuous video-based monitoring in enhancing the reliability of CAD risk assessment and supports the design of DigitalShadow as a longitudinal, real-time health monitoring system.

\subsection{LiveCAD learns clinically relevant facial features associated with CAD}
To gain insight into the facial features that LiveCAD associates with CAD, we employed Gradient-weighted Class Activation Mapping (Grad-CAM) for model interpretability (Methods). This technique allows visualization of the regions in facial images that most influence the model’s predictions. As illustrated in Figure \ref{fig2}d, LiveCAD consistently attends to several facial regions previously reported in clinical literature to be associated with CAD, including earlobe creases, malar (cheek) pouches, and nasolabial folds. These localized heatmaps indicate that LiveCAD is not relying on arbitrary visual cues but is instead focusing on medically plausible markers. Such interpretability not only supports the clinical relevance and transparency of the model’s predictions but also enhances trust and usability in real-world healthcare settings. These visualizations provide compelling evidence that LiveCAD is capable of learning meaningful, domain-specific facial biomarkers related to cardiovascular health.

\subsection{DigitalShadow functions as a 24/7 family doctor for continuous CAD risk monitoring from live video streams}
As LiveCAD processes longer durations of video, it analyzes a sequence of facial frames to produce a corresponding time series of CAD risk scores, enabling real-time and longitudinal health monitoring. Figure \ref{fig2}e illustrates the distribution of CAD risk values over time for two representative patients (IDs: 622022 and 52453625), while Figure \ref{fig2}f presents the aggregated distributions of CAD risk scores for all healthy subjects and CAD patients, in which both distributions show significant divergence (KL divergence = 7.47). Based on this, by calculating the difference in the distribution of CAD risk values between two time periods and the corresponding P-value, it is possible to statistically determine whether there is an overall increase or decrease in CAD risk.

Importantly, this video-based approach enables temporal risk profiling, capturing fluctuations and trends that are not observable from single images. As shown in Figure \ref{fig2}g, longer observation durations (i.e., risk predictions aggregated over more video frames) lead to significantly increased stability and robustness in average CAD risk estimates. In repeated experiments (n = 100), the variance of risk scores decreases with longer observation windows, confirming the system’s enhanced reliability in dynamic, real-world scenarios. These findings highlight a key insight that relying on one or a few facial images is insufficient for accurate CAD prediction. Instead, LiveCAD’s continuous video analysis enables comprehensive, high-resolution risk assessment, positioning DigitalShadow as a 24/7 virtual family doctor, which is capable of delivering real-time, passive, and privacy-preserving health surveillance for early intervention and personalized care.

\subsection{DigitalShadow generates user-friendly CAD risk reports}
DigitalShadow integrates LiveCAD's CAD risk assessments with personalized health records and leverages a LLM to reason and generate user-friendly risk reports. By combining CAD predictions with context-specific health data, DigitalShadow delivers insightful, easy-to-understand summaries for users and healthcare providers alike. Figure \ref{fig3} presents two illustrative cases, showcasing DigitalShadow’s ability to generate risk reports. The primary distinctions between the two cases lie in the chief complaint information retrieved from the personalized health database and the CAD risk predictions generated by LiveCAD over time. These varying data inputs are analyzed by DigitalShadow, which then synthesizes a comprehensive assessment of the user's CAD risk level.

DigitalShadow not only provides an accurate risk level but also offers tailored health recommendations based on the individual’s unique health profile and historical risk data. This process emphasizes DigitalShadow's capacity to deliver actionable, personalized insights that guide users toward informed health decisions. Ultimately, DigitalShadow showcases its ability to generate clear and actionable health reports, making complex CAD risk information easily accessible to users without specialized medical knowledge.

\subsection{DigitalShadow is computationally efficient and time saving}
The traditional process for diagnosing CAD typically involves multiple sequential stages, including initial assessment, risk stratification, diagnostic testing, and clinical diagnosis. Initially, the patient consults a primary care physician or cardiologist, who performs a comprehensive evaluation based on medical history, presenting symptoms, and a physical examination. Key risk factors such as age, sex, family history, smoking status, hypertension, hyperlipidemia, and diabetes are carefully assessed using validated scoring tools, such as the Framingham Risk Score or ASCVD risk estimator. Based on these risk factors and symptoms, the physician may order various diagnostic tests, such as an electrocardiogram (ECG), stress test, echocardiogram, or coronary angiography, to assess heart function and blood flow. The overall diagnostic pathway from the initial clinical visit to the final confirmed diagnosis of CAD could take a few weeks, depending on appointment availability, test complexity, and institutional workflow. This delay may impact timely intervention and disease management, especially in asymptomatic or early-stage cases.

In contrast, DigitalShadow acts as an early warning system that passively and contactlessly monitors facial features to infer CAD risk, delivering preliminary results within a few seconds as shown in \textbf{Table} \ref{table_profile1} and \ref{table_profile2}. This dramatic reduction in diagnostic latency highlights DigitalShadow’s efficiency and immediacy, allowing for earlier detection and intervention than traditional methods, particularly in asymptomatic populations or remote care settings.

To ensure practicality in real-world deployment, DigitalShadow is optimized for edge computing. It could be deployed on the Raspberry Pi 4 Model B, a compact, affordable computing device, capable of supporting the system’s computational demands. The camera connects via ribbon cable for seamless real-time video streaming. This hardware-software integration supports low-latency processing, enabling real-time facial analysis and CAD risk prediction entirely on the edge without uploading sensitive data to external servers. Due to its low cost, compact size, and robust performance, this setup is highly scalable and ideal for diverse deployment settings, such as home-based eldercare, remote clinics, or single-room hospital surveillance. Together with its privacy-preserving design, DigitalShadow provides a practical and intelligent solution for continuous CAD risk monitoring in a wide range of real-world healthcare environments.

\begin{table*}[!htb]
    \centering
        \caption{Resource usage and inference performance of DigitalShadow.}\label{table_profile1}
    \begin{tabular}{c|c}
\toprule
Resource & Value \\
\toprule
GPU Memory Usage (Inference) & 1.73 GB \\
RAM Usage (Inference) & 1.20 GB \\
Inference Time per Frame (Dev Machine, GPU) & 1.7 seconds \\
Inference Time per Frame (Edge Device, GPU) & 2.1 seconds \\
Inference Time per Frame (Edge Device, CPU) & 7.8 seconds \\
\toprule
\end{tabular}
\end{table*}

\begin{table}[!htb]
    \centering
        \caption{Inference efficiency of DigitalShadow across different computational devices.}\label{table_profile2}
    \begin{tabular}{c|c}
\toprule
Device & Inference Time per Frame (s)\\
\toprule
CPU & 7.8$\pm$2.2 \\
A100 (80GB) & 1.7$\pm$0.1 \\
A100 (40GB) & 1.7$\pm$0.1 \\
V100 & 1.8$\pm$0.1 \\
RTX 2080 & 1.9$\pm$0.2 \\
Quadro P6000 & 2$\pm$0.1 \\
Tesla P100 & 2$\pm$0.5 \\
GTX 1080 & 2.1$\pm$0.2 \\
\toprule
\end{tabular}
\end{table}

\section{Methods}
\subsection{Ethics approval}
This study utilized advanced deep learning techniques trained on a combination of publicly available facial image datasets and a proprietary dataset comprising facial images from clinically diagnosed CAD patients. All procedures were conducted in strict accordance with internationally recognized ethical standards for biomedical research involving human subjects. Ethical approval for the study was obtained from the Ethics Committee of Beijing Anzhen Hospital, affiliated with Capital Medical University, Beijing, China, under reference number (2022) Ethical Review of Science and Technology No. 38. In addition, the study was reviewed and approved by the Institutional Biosafety and Bioethics Committee (IBEC) at King Abdullah University of Science and Technology (KAUST), under protocol number 23IBEC100.

For the use of publicly available facial image datasets, we rigorously adhered to the specific usage policies and licensing agreements associated with each dataset. As these datasets are openly released for research purposes, no additional informed consent was required for their use. In contrast, for the in-house clinical dataset, all participants were recruited under protocols approved by the respective institutional review boards, and written informed consent was obtained from every individual whose data was included in the study. These participants were fully informed about the study’s purpose, data handling practices, and privacy safeguards. To further ensure data protection and privacy, all in-house facial images were anonymized prior to analysis, in compliance with data protection regulations such as the General Data Protection Regulation (GDPR) and China’s Personal Information Protection Law. The ethical rigor maintained throughout this study underscores the responsible and human-centered approach to AI development and deployment in healthcare.

\subsection{Study design and population}
This study was conducted between December 2022 and March 2025, focusing on the recruitment of CAD patients and healthy subjects for the development and clinical validation of the DigitalShadow solution. 

The CAD patient cohort consisted of critically ill patients aged between 18 and 85 years who were treated at participating hospitals. This included patients who underwent emergency assessment, observation, resuscitation, or were admitted to the Emergency Intensive Care Unit (EICU). Patients included in the CAD patient cohort were diagnosed with CAD based on findings from either coronary angiography or coronary computed tomography angiography (CCTA). Exclusion criteria for this group included:

1. A history of percutaneous coronary intervention (PCI)

2. A history of coronary artery bypass grafting (CABG)

3. A history of major structural heart diseases, such as congenital heart disease, valvular heart disease, or large-vessel cardiovascular anomalies

The control group included subjects aged between 18 and 85 years during the same period. These subjects had no significant changes on electrocardiograms (ECGs) within the past six months and no documented history of coronary artery disease, diabetes mellitus, hypertension, or other major chronic comorbidities. 

The LiveCAD model was developed through a two-stage training pipeline: pre-training and fine-tuning. During the pre-training stage, the facial foundation model was pre-trained using a large-scale dataset of 21 million publicly available facial images, representing approximately 1 million unique identities, with a resolution of 448×448 pixels. The dataset was aggregated from sources including WebFace and Arc2Face, providing a diverse representation of facial structures and demographics. During the fine-tuning stage, the model was then fine-tuned for CAD risk assessment using a curated in-house clinical dataset consisting of 7,004 facial images from 1,478 CAD patients and 273 healthy subjects. These data were collected from four collaborating hospitals across China, including Beijing Anzhen Hospital, Beijing Changping Hospital, Tongji Hospital and Daqing Longnan Hospital in China. Each subject contributed up to four standardized facial images from different angles (front, left, right, and top views), ensuring multi-perspective coverage of facial features. Additionally, video data from a subset of CAD patients was recorded at Beijing Anzhen Hospital to enable temporal risk analysis and longitudinal monitoring.

\subsection{Video processing and face registration}
DigitalShadow leverages standard home surveillance cameras to continuously capture real-time video streams for passive health monitoring (\textbf{Algorithm \ref{algo_ds}}). For this purpose, we employed the Hikvision DS-E12 USB camera, which supports a frame rate of 30 frames per second (FPS) at a 1920×1080 pixel resolution (1080p), enabling high-definition video capture suitable for detailed facial analysis. The real-time video data is processed using a software pipeline built with Python 3.10 \cite{10.5555/1593511}, mmcv 2.1.0 \cite{mmcv}, and PyTorch 1.13.1 \cite{paszke2019pytorch}. After detection, only facial frames corresponding to the pre-registered user are cropped and passed to the LiveCAD module for CAD risk assessment. The predicted results are then stored in the user's personalized health database, which supports longitudinal monitoring and subsequent report generation (\textbf{Algorithm \ref{algo_Face_recognition}}). This modular and privacy-conscious pipeline ensures that only authorized users are monitored, and only the relevant facial data is used for health analysis, maintaining both accuracy and data security in a home setting.

\begin{algorithm}
\caption{DigitalShadow: Personalized Real-Time Risk Analysis via Visual Observation}\label{algo_ds}
\begin{algorithmic}[1]
\Require Live video stream $V$, personalized database $D$, temporal analysis window size $T$
\For{each frame $v_t$ at timestamp $t$ in $V$}
    \State \textbf{Face Detection:} Extract set of detected faces $F_t \gets \textsc{DetectFaces}(v_t)$
    \For{each face $f_i \in F_t$}
        \State \textbf{Identification:} Match $f_i$ against known user profiles via $\textsc{FaceRecognition}(f_i)$
        \If{match found}
            \State Let $f_t^{user} \gets f_i$
        \EndIf
    \EndFor
    \If{$f_t^{user}$ exists}
        \State \textbf{Risk Evaluation:} Compute instantaneous CAD risk score $r_t \gets \textsc{AssessRisk}(f_t^{user})$ with LiveCAD
        \State Store $\langle t, r_t \rangle$ into database $D$
    \EndIf
    \State \textbf{Temporal Aggregation:} For the current sliding window $T_t$, compute statistical descriptors (mean, variance, distribution) from $D$
    \State \textbf{Change Detection:} Compare statistical patterns of $T_t$ and $T_{t-1}$ to identify shifts in CAD risk trends
    \State \textbf{High-Level Analysis:} Invoke a Large Language Model $\textit{LLM}$ to reason over detected changes and summarize insights
\EndFor
\end{algorithmic}
\end{algorithm}

\begin{algorithm}
\caption{FaceRecognition: Identity Matching via Embedding Similarity}\label{algo_Face_recognition}
\begin{algorithmic}[1]
\Require Cropped face image $f_c$ (from video), labeled face database $F$, personalized database $D$
\State \textbf{Embedding:} Generate feature embedding $e_c \gets \textsc{Embed}(f_c)$
\State Embed all faces in $F$: $E = \{ e_i \gets \textsc{Embed}(f_i) \mid f_i \in F \}$
\State Initialize $d_{\min} \gets \infty$, $f^* \gets \emptyset$
\For{each $e_i \in E$}
    \State Compute distance $d_i \gets \|\mathbf{e}_i - \mathbf{e}_c\|_2$
    \If{$d_i < d_{\min}$}
        \State $d_{\min} \gets d_i$
        \State $f^* \gets f_i$
    \EndIf
\EndFor
\If{$f^* \neq \emptyset$}
    \State Assign identity $\textsc{Label}(f^*)$ to $f_c$
    \State Register $(f_c, \textsc{Label}(f^*))$ into personalized database $D$
\EndIf
\end{algorithmic}
\end{algorithm}

\subsection{Pre-train facial foundational model}
\label{sec:pretrain_model}
During pre-training, we leverage the BEiT v2~\cite{peng2022beit} framework to pre-train a foundational facial model using 21 million high-resolution facial images (448$\times$448) of 1 million distinct identities sourced from the WebFace~\cite{zhu2021webface260m} and Arc2Face~\cite{papantoniou2024arc2face} datasets. This pre-training is motivated by the need for robust, semantic-level feature extraction capable of capturing intricate facial features associated with early indicators of CAD. By moving beyond pixel-level information, BEiT v2 enables a more powerful extraction of semantically rich features that are crucial for identifying subtle facial indicators. Masked image modeling has demonstrated strong potential in self-supervised learning by recovering masked image patches, facilitating the extraction of high-level semantic features~\cite{xie2022simmim}. BEiT v2 extends this concept by introducing Vector-Quantized Knowledge Distillation (VQ-KD), which promotes the modeling process from pixel level to semantic level. This transition is particularly useful for facial image analysis, where semantic features related to early CAD indicators are crucial.

\textbf{Encoder for Image Representation}\quad
The input images \( x \in \mathbb{R}^{448 \times 448 \times 3} \) are split into patches of size \( 16 \times 16 \), resulting in \( N = 196 \) patches per image. Each patch is linearly projected into a higher-dimensional feature space, forming patch embeddings \( h_i \in \mathbb{R}^{768} \), which are then processed by the Vision Transformer (ViT) encoder~\cite{dosovitskiy2020image} to produce the final representation \( H \):
\begin{equation}
H = \text{ViT}([\text{Linear}(x_1), \text{Linear}(x_2), \dots, \text{Linear}(x_N)])
\end{equation}
where \( H=[h_1, h_2, \dots, h_N] \) represents the encoded patch sequence after applying the transformer layers in ViT.

\textbf{Vector-Quantized Knowledge Distillation} \quad
The VQ-KD mechanism quantizes the continuous feature representations into discrete tokens that serve as semantic representations of the image patches. The codebook \( V \in \mathbb{R}^{K \times D} \) contains \( K = 8192 \) discrete codebook embeddings, where each embedding has a dimension \( D = 32 \). The quantization of each patch embedding \( h_i \) is defined as:
\begin{equation}
z_i = \arg\min_{j} \|\ell_2(h_i) - \ell_2(v_j)\|^2 \quad \forall j \in [1, K]
\end{equation}
where \( \ell_2(\cdot) \) represents the \( \ell_2 \)-normalization for the codebook lookup. This quantization ensures that each facial image is represented by a compact set of visual tokens, capturing high-level semantics essential for CAD detection.

\textbf{Decoder in BEiT v2}\quad
The decoder plays a crucial role in reconstructing the semantic features from the discrete visual tokens. After the input image patches are quantized into discrete tokens using the visual tokenizer, these tokens are fed into a Transformer-based decoder. The decoder’s objective is to generate semantic representations that closely match the output of a teacher model, such as CLIP~\cite{radford2021learning} or DINO~\cite{caron2021emerging}, which guides the semantic content of the image patches. Specifically, given the sequence of \( N \) quantized tokens \( \{v_{z_1}, v_{z_2}, \dots, v_{z_N}\} \), where each token corresponds to an image patch, the decoder processes this sequence through several Transformer layers to produce output vectors \( \{o_1, o_2, \dots, o_N\} \). The goal of the decoder is to maximize the similarity between these output vectors \( o_i \) and the corresponding semantic vectors \( t_i \) provided by the teacher model.

\textbf{Overall Training Loss for VQ-KD} The training objective of the VQ-KD tokenizer is designed to recover the semantic features from the teacher model by maximizing the cosine similarity between the output vectors \( o_i \) and the teacher's guidance \( t_i \). The overall loss function for VQ-KD is composed of three terms:
\begin{align}
\mathcal{L}_{\text{VQ-KD}} = \sum_{x \in D} \sum_{i=1}^{N} & \left[ \text{cos}(o_i, t_i) 
    - \|\text{sg}[\ell_2(h_i)] - \ell_2(v_{z_i})\|^2 \right. \nonumber \\
    & \left. - \|\ell_2(h_i) - \text{sg}[\ell_2(v_{z_i})]\|^2 \right]
\end{align}
Here, Cosine similarity \( \text{cos}(o_i, t_i) \) measures how similar the output of the model \( o_i \) is to the semantic guidance provided by the teacher model \( t_i \) for each patch \( i \). Cosine similarity ranges from -1 (most dissimilar) to 1 (most similar), and the goal is to maximize this similarity. This ensures that the model’s output is semantically aligned with the teacher model's representation, meaning that the learned visual tokens represent the high-level features of the image patches. The first regularization term \( \|\text{sg}[\ell_2(h_i)] - \ell_2(v_{z_i})\|^2 \) ensures that the quantized codebook vector \( v_{z_i} \) accurately represents the original patch embedding \( h_i \). The \( \ell_2 \)-normalization of both \( h_i \) and \( v_{z_i} \) ensures that these vectors are normalized to unit length, which stabilizes the training. The stop-gradient operator \( \text{sg}[\cdot] \) prevents updates to the patch embedding \( h_i \) based on this term, allowing only the codebook vector \( v_{z_i} \) to be updated to better represent \( h_i \). The second regularization term \( \|\ell_2(h_i) - \text{sg}[\ell_2(v_{z_i})]\|^2 \) works in the opposite direction: it ensures that the patch embedding \( h_i \) is updated to be closer to the quantized vector \( v_{z_i} \), while preventing updates to the codebook vector \( v_{z_i} \). Thus, the loss function encourages the decoder to produce semantic representations that match the teacher model while maintaining a stable and diverse set of visual tokens in the codebook.

\textbf{Masked Image Modeling for Pre-training} \quad
Once the tokenizer is trained, the facial foundational model is pretrained using Masked Image Modeling (MIM), where approximately 40\% of the image patches are randomly masked. The model is tasked with recovering the original visual tokens for the masked patches. For each masked patch \( i \in M \), the pretraining objective is defined as:
\begin{equation}
\mathcal{L}_{\text{MIM}} = - \sum_{x \in D} \sum_{i \in M} \log p(z_i | h_i)
\end{equation}
where \( p(z_i | h_i) \) represents the predicted probability of the correct visual token \( z_i \) based on the corrupted input. This objective forces the model to learn meaningful global representations by recovering semantic tokens, which are crucial for detecting subtle facial features linked to CAD.

\textbf{Patch Aggregation Strategy} \quad
To improve global image representations, a patch 
aggregation strategy that encourages the [CLS] token to aggregate information from all patches is introduced. The [CLS] token is a special token that represents the entire image. By aggregating the information from all patches into this token, the model can better capture global context, which is essential for tasks requiring comprehensive understanding of the image, such as early detection of CAD.

The patch aggregation strategy is achieved by adding an additional MIM loss on the [CLS] token. The final training loss is the sum of the MIM loss and this additional patch aggregation loss, which is defined as:
\begin{equation}
\mathcal{L} = \mathcal{L}_{\text{MIM}} + \mathcal{L}_{\text{CLS}}
\end{equation}
where \( \mathcal{L}_{\text{CLS}} \) is the loss that ensures the [CLS] token aggregates the information from all patches effectively.

The combination of these two losses encourages the model not only to recover the masked patches but also to create a strong global representation that captures the full context of the image. This is particularly important for detecting subtle facial features associated with CAD.

\textbf{Training Setup} \quad
The pre-training process required approximately seven days on a cluster of eight NVIDIA A100 GPUs (80GB). The model was trained with an AdamW optimizer using a peak learning rate of 1.5e-3 and a weight decay of 0.05. The input resolution of the images was set to 448$\times$448, and a batch size of 2,048 was used. The model's large-scale training on diverse facial images enables it to generalize well across various facial types and image conditions, making it highly suitable for tasks like early CAD detection. By training on this large and diverse dataset, the pre-trained model becomes a powerful tool for embedding facial features. The next step involves fine-tuning this model on a specialized CAD dataset for personalized risk assessment.



\subsection{Fine-tune model for CAD classification}

The pre-trained facial foundational model was further fine-tuned using a dataset of in-house facial images collected from a diverse cohort of individuals, comprising 1478 CAD patients and 273 healthy individuals, sourced from four hospitals across China. To ensure rigorous evaluation and mitigate potential biases, the dataset was carefully divided into training and testing subsets based on patient identity information. The fine-tuned model is called LiveCAD. To validate the model’s performance, we employed a five-fold cross-validation approach. This method involved partitioning the data into five distinct folds, with the model being iteratively trained and tested on different folds. Each fold served as both a training and testing set at various stages, which helped to assess the model's robustness, reliability, and generalizability while minimizing overfitting. The entire fine-tuning process required approximately 24 hours to complete. This computation was performed using a single NVIDIA A100 (80GB) GPU, operating on an Ubuntu 20.04 system with Python 3.10, PyTorch 2.1, and CUDA 12.2. The extensive computational resources and processing time reflect the model’s complexity and the scale of the dataset, highlighting the thoroughness of the fine-tuning procedure.

\subsection{Evaluation metrics}\label{metrics}

We used accuracy, precision, recall and f1-score to measure the performance of LiveCAD as below:

\begin{equation}
    Accuracy = \frac{TP+TN}{TP+TN+FP+FN}
\end{equation}

\begin{equation}
    Precision = \frac{TP}{TP+FP}
\end{equation}

\begin{equation}
    Recall = \frac{TP}{TP+TN}
\end{equation}

\begin{equation}
    F1-score = \frac{2*precision*recall}{precision + recall}
\end{equation}

\noindent
where TP, TN, FP, and FN stand for true positive, true negative, false positive and false negative.

\subsection{DigitalShadow Agent with large language model}
We utilized LangChain and DeepSeek-R1 7B\cite{guo2025deepseek} to customize the DigitalShadow Agent for handling results from LiveCAD and integrating personalized health records. This customization enables the DigitalShadow Agent to analyze complex data and generate structured reports effectively. To facilitate this process, we designed specific prompts and templates for the language model to ensure accurate and coherent output. Details of these prompts are included in the supplementary information. The integration of these advanced language models enhances the DigitalShadow Agent’s capability to provide comprehensive and actionable insights based on the CAD risk assessments and individual health records.

\subsection{Model visualization}
To better understand the facial features associated with CAD that our trained model can capture, we employed Gradient-weighted Class Activation Mapping (Grad-CAM). Grad-CAM is a powerful technique used to generate visual explanations of a model’s decision-making process by highlighting the regions of an image that contribute most significantly to the final prediction. This method works by backpropagating the gradients of the output score for CAD with respect to the activations of a selected layer within the convolutional neural network. By doing so, Grad-CAM computes the relative importance of each pixel or region in the input image, allowing us to visually identify which facial features the model relies on when assessing CAD risk. This technique provides valuable insights into the model's behavior and helps in interpreting its decision-making process, ensuring that the model’s reasoning aligns with clinically relevant facial characteristics associated with CAD.

Given an image \( I \) as input, it is divided into non-overlapping patches \( P \), where each patch \( p_i \) corresponds to a portion of the image. The ViT model maps these patches to a sequence of embeddings using a patch embedding layer. Then, the model predicts a class score \( y^c \) for class \( c \). The score \( y^c \) is defined as:
    \begin{equation}
    y^c = f(I)
    \end{equation}
where \( f \) is the output of the model for the class \( c \).
    
The patch embeddings are processed through several transformer layers. Grad-CAM can be applied to any of the attention layers, but we typically focus on the final attention layer, where the class token attends to all the patches. The attention weights represent the contribution of each patch to the final prediction. The class token output \( y^c \) is the model’s prediction for class \( c \). We compute the gradient of \( y^c \) with respect to the patch embeddings \( E^p \), which are the activations of the patches after the final attention layer:
\begin{equation}
\frac{\partial y^c}{\partial E^p}
\end{equation}

the global average pooling of these gradients over all patches to obtain the weights \( \alpha_p^c \) for each patch embedding \( E^p \):
\begin{equation}
\alpha_p^c = \frac{1}{Z_p} \sum_{i,j} \frac{\partial y^c}{\partial E_{i,j}^p}
\end{equation}
where \( Z_p \) is the number of patches. The final localization map is then computed as:
\begin{equation}
L^c_{\text{ViT-Grad-CAM}} = \text{ReLU}\left( \sum_p \alpha_p^c E^p \right)
\end{equation}
The resulting localization map \( L^c_{\text{ViT-Grad-CAM}} \) highlights the patches that are most relevant to the class \( c \), thus providing insights into the decision-making process of the Vision Transformer.

\section{Discussion}
DigitalShadow represents a significant advancement in the early detection of CAD, offering a non-invasive, contactless, and intelligent risk assessment system. By leveraging a vertically fine-tuned facial foundation model, DigitalShadow holds promise for improving the accessibility, efficiency, and scalability of CAD screening. Its ability to passively extract risk-related facial features from live video streams without active user participation makes it especially valuable for large-scale population screening, telemedicine, and home-based health monitoring.

Despite its clear advantages, several critical areas warrant further exploration to maximize DigitalShadow’s clinical impact and generalizability. A key limitation lies in the potential demographic bias arising from the composition of the training data. As the current model has been fine-tuned predominantly on a Chinese population, its performance may degrade when applied to individuals from diverse ethnic and geographic backgrounds. To address this, future research should aim to expand the training dataset to include multi-ethnic facial images from a variety of regions, ensuring fairness, inclusiveness, and robustness in CAD risk assessment across global populations.

Furthermore, facial features alone may not sufficiently capture the full spectrum of an individual’s CAD risk. The integration of additional physiological and behavioral markers, such as voice biomarkers, retinal imaging, heart rate variability, and lifestyle factors, could substantially enhance the system's predictive accuracy. Future iterations of DigitalShadow could leverage multi-modal data fusion and cross-modal learning techniques to enable more comprehensive and personalized assessments.

Given the inherently sensitive nature of medical data, data privacy and security remain paramount. While DigitalShadow supports local deployment to reduce risks associated with centralized data processing, further efforts are required to incorporate privacy-preserving machine learning frameworks, such as differential privacy, federated learning, and homomorphic encryption. Additionally, enhancing the transparency and interpretability of DigitalShadow’s AI models will be essential to meet regulatory requirements, foster clinician trust, and promote user adoption.

By addressing these limitations and pursuing these future directions, DigitalShadow can evolve into a trusted and indispensable tool in AI-driven preventive cardiology, enabling early detection, continuous monitoring, and timely intervention for CAD on a global scale.

\section{Acknowledgements}

\textbf{Special Thanks: }We thank all the participants and the institutions for supporting this study.\\

\noindent
\textbf{Funding: }J.Z., Z.H., G.L., W.H., Y.C., X.G. were supported by the King Abdullah University of Science and Technology (KAUST) Office of Research Administration (ORA) under Award No REI/1/5234-01-01, REI/1/5414-01-01, REI/1/5289-01-01, REI/1/5404-01-01, REI/1/5992-01-01, URF/1/4663-01-01, Center of Excellence for Smart Health (KCSH), under award number 5932, and Center of Excellence on Generative AI, under award number 5940. X.H. was supported by National Natural Science Foundation of China (62272327, 12371527), the Clinical Medicine Development Special Project "Yangfan 3.0" of Beijing Municipal Hospital Administration Center(YGLX202), and CSC Clinical Research Special Fund of the Chinese Medical Association in 2024(CSCF2024B03).\\

\noindent
\textbf{Competing Interests: }The authors have declared no competing interests.\\

\noindent
\textbf{Credit author statement: }J.Z., X.G. contributed to the concept of the study and designed the research. J.X., X.G., X.H., M.X., X.H., G.W., X.L., J.C., X.B., B.W., J.S., X.W., W.D., Z.G., L.B., Y.P. collected and processed the data. J.Z., X.G. conducted the study. J.Z. analyzed the data. J.Z., Z.H., X.G. co-wrote the manuscript. J.Z., Z.H., X.G. critically revised the manuscript. J.Z., Z.H., G.L., W.H., X.G. performed the technical review. All authors discussed the results and provided comments regarding the manuscript.\\

\noindent
\textbf{Data availability: }The data supporting the findings of this study are categorized into two groups: publicly available datasets and restricted-access datasets. Public datasets are cited within the manuscript and referenced accordingly. Due to privacy regulations, the patient data collected and analyzed in this study cannot be made publicly available. These data were approved for use exclusively within Anzhen Hospital for research purposes. Any external access or reuse of this data requires additional informed consent from participants and formal ethical approval from the Anzhen Hospital Institutional Review Board. Data access may be granted upon reasonable request to the corresponding author, subject to these conditions.\\

\noindent
\textbf{Code availability: }To promote academic exchanges, under the framework of data and privacy security, the inference code proposed by DigitalShadow is publicly available at https://github.com/JoshuaChou2018/DigitalShadow, and the trained model is not released due to privacy issues. In the case of noncommercial use, access to the latest trained model weights may be granted upon reasonable request to the corresponding author, subject to these conditions. We have also provided comprehensive details of our methodology and experimental procedures in the main text and Supplementary Information, enabling reproducibility of the experiments. Several core components utilized in our work are open-source and publicly available, including PyTorch (v2.1), mmcv (v2.1.0), and langchain\_community (v0.3.20). 

{
\bibliographystyle{IEEEtran}
\bibliography{reg}
}

\end{document}


%
\title{Supplementary Information for A Privacy-Preserving Elderly Healthcare System for Early Warning of Critical Diseases}
%
%
%
%

\author{Juexiao~Zhou$^{1,2}$, Xin~Gao$^{1,2,*}$
\thanks{
$^1$Computer Science Program, Computer, Electrical and Mathematical Sciences and Engineering Division, King Abdullah University of Science and Technology (KAUST), Thuwal 23955-6900, Kingdom of Saudi Arabia\\
$^2$Computational Bioscience Research Center, King Abdullah University of Science and Technology, Thuwal 23955-6900, Kingdom of Saudi Arabia\\
$^*$Corresponding author. e-mail: xin.gao@kaust.edu.sa\\
$^\#$These authors contributed equally.
}}

%
%

\markboth{}%
\maketitle

\IEEEdisplaynontitleabstractindextext

%
\IEEEpeerreviewmaketitle

\section{Introduction}
this is a minor point as most people would not even notice if the said evil this is a minor point as most people would not even notice if the said evil this is a minor point as most people would not even notice if the said evil this is a minor point as most people would not even notice if the said evil this is a minor point as most people would not even notice if the said evil

{
\bibliographystyle{IEEEtran}
\bibliography{reg}
}